\RequirePackage{fix-cm}
\documentclass[twocolumn]{svjour3}          % twocolumn
\smartqed  % flush right qed marks, e.g. at end of proof
\usepackage{graphicx}
\usepackage{amssymb}
\usepackage{hyperref}
\usepackage{amsmath}
\usepackage{color}
\usepackage{booktabs}
\usepackage[table,xcdraw]{xcolor}
\usepackage{algorithmic}
\usepackage{algorithm}
\usepackage{lipsum}
\usepackage{epstopdf}
\usepackage{multirow}
\usepackage{ulem}
\usepackage{subfig}
\newtheorem{thm}{Theorem}

\usepackage[misc]{ifsym}
\usepackage{bbding}

\begin{document}

\title{A Study on the Uncertainty of Convolutional Layers in Deep Neural Networks %\thanks{Grants or other notes
%about the article that should go on the front page should be
%placed here. General acknowledgments should be placed at the end of the article.}
}
%\subtitle{Do you have a subtitle?\\ If so, write it here}

%\titlerunning{Short form of title}        % if too long for running head

\author{Haojing Shen\textsuperscript{1}  \and
        Sihong Chen\textsuperscript{1} \and
        Ran Wang\textsuperscript{2,*} 
%        Xizhao Wang\textsuperscript{1,*}
}
%\author{First author  \and
%	Second author \and
%}

%\authorrunning{Short form of author list} % if too long for running head

\institute{\Letter Ran Wang \\
	\email{wangran@szu.edu.cn} \\
	 \at
            {1} Big Data Institute, College of Computer Science and Software Engineering, Guangdong Key Lab. of Intelligent Information Processing, Shenzhen University, Shenzhen 518060, Guangdong, China \\
%              \email{winddyakoky@gmail.com;2651713361@qq.com.}           %  \\
%             \emph{Present address:} of F. Author  %  if needed
           \and
     \at
            {2} The College of Mathematics and Statistics, Shenzhen University, Shenzhen 518060, China and also with the Shenzhen Key Laboratory of Advanced Machine Learning and Applications, Shenzhen University, Shenzhen 518060, China. \\
%              \email{wangran@szu.edu.cn}
}

\date{Received: date / Accepted: date}
% The correct dates will be entered by the editor

\maketitle

\begin{abstract}

%This paper shows a Min-Max property existing in connection weights of convolutional layers of a neural network structure, i.e., the LeNet. Specifically, the Min-Max property means that, during the process of BP-based training an LeNet, the weight values of convolutional layers are becoming far away from their centers of intervals, i.e., decreasing to their minimum or increasing to their maximum. From perspective of uncertainty, we demonstrate that the Min-Max property corresponds to minimizing the fuzziness of model parameters through a simplified formulation of convolution. It is experimentally confirmed that the model with Min-Max property will have a better adversarial robustness, and incorporating the Min-Max property into the loss function design can improve the adversarial robustness. The paper points out a changing tend of uncertainty at convolutional layers in LeNet and gives some insights to the interpretability of convolution.
This paper shows a Min-Max property existing in the connection weights of the convolutional layers in a neural network structure, i.e., the LeNet. Specifically, the Min-Max property means that, during the back propagation-based training for LeNet, the weights of the convolutional layers will become far away from their centers of intervals, i.e., decreasing to their minimum or increasing to their maximum. From the perspective of uncertainty, we demonstrate that the Min-Max property corresponds to minimizing the fuzziness of the model parameters through a simplified formulation of convolution. It is experimentally confirmed that the model with the Min-Max property has a stronger adversarial robustness, thus this property can be incorporated into the design of loss function. This paper points out a changing tendency of uncertainty in the convolutional layers of LeNet structure, and gives some insights to the interpretability of convolution.

% \PACS{PACS code1 \and PACS code2 \and more}
% \subclass{MSC code1 \and MSC code2 \and more}
\end{abstract}

\section{Introduction}
\label{intro}
Deep neural networks (DNNs) are vulnerable to adversarial examples \cite{r21_szegedy2013intriguing}, thus improving adversarial robustness is a critical issue for DNN models. On the one hand, adversary can construct various attacks against the model, including white-box attacks \cite{r21_szegedy2013intriguing,r22_madry2017towards,r23_goodfellow2014explaining,r24_carlini2017towards,r26_athalye2018obfuscated,r27_moosavi2016deepfool,r28_papernot2016limitations} and black-box attacks \cite{r31_liu2016delving,r59_chen2017zoo,r60_su2019one,r61_zhao2017generating}. On the other hand, the model can adopt certain defensive methods to improve the adversarial robustness, such as distillation training \cite{r29_papernot2016distillation}, normalization \cite{r47_qin2019adversarial}, and adversarial training \cite{r23_goodfellow2014explaining,r20_kurakin2018ensemble,r25_abbasi2017robustness,r22_madry2017towards,r18_hein2017formal,r19_sinha2017certifying}. The game between adversarial attacks and adversarial defenses becomes more and more intense.

Uncertainty is a natural phenomenon in machine learning. It is pervasive in data and models, which can be embedded into many processes like data preparation, learning, and reasoning. Usually, uncertainty can be estimated by entropy \cite{r64_shannon2001mathematical} or fuzziness \cite{r65_kosko1990fuzziness}. In order to improve the adversarial robustness of a DNN model, the relationship between uncertainty and adversarial robustness has been studied \cite{r66_Liu_2019_ICCV,r67_bradshaw2017adversarial}. For example, some works try to improve the adversarial robustness by minimizing the incorrect-class entropy \cite{r62_chen2019improving} or realize the adversarial example detection by applying the concept of mutual information \cite{r63_smith2018understanding}. 

%However, the above-mentioned works mainly focus on the uncertainty of the model's prediction or data. In this paper, we study the uncertainty of parameters of convolutional layers by taking LeNet as an example. We propose a Min-Max property that makes the value of parameters in convolutions either further close to zero or further far away from zero. In essence, this is an off-center phenomenon which means that the fuzziness of convolutional layers is going to be small gradually. We theoretically prove that the model with Min-Max property has better adversarial robustness than the standard training model. Moreover, from the perspective of uncertainty, we measure the uncertainty of parameters of convolutional layers using fuzziness. We point out that minimizing our objective function, which is designed according to Min-Max property, equals minimizing the uncertainty of parameters of convolutional layers. It is experimentally confirmed that the model with Min-Max the property will have better adversarial robustness, and incorporating the Min-Max property into the loss function design can improve the adversarial robustness. The paper points out a changing tendency of uncertainty in convolutional layers in LeNet and gives some insights into the interpretability of convolution.

It is noteworthy that the above-mentioned works mainly focus on the uncertainty of the model's prediction or data. In this paper, we study it from a new perspective, i.e., the uncertainty of parameters in the convolutional layers of a DNN model. By taking LeNet as an example, we propose a Min-Max property that makes the parameters in the convolutional layers far away from their centers of intervals, i.e., the parameters close to zero will be closer to zero, and the parameters far away from zero will be further away from zero. Essentially, this is an off-center phenomenon that the fuzziness of the convolutional layers is decreasing gradually to be small. We theoretically prove that the model with the Min-Max property has stronger adversarial robustness than the standard model. Moreover, we point out that minimizing our objective function, which is designed according to the Min-Max property, equals to minimizing the uncertainty of the parameters in the convolutional layers. Experimentally, it is also confirmed that the model with the Min-Max property has stronger adversarial robustness, thus this property can be incorporated into the design of loss function for DNNs. This paper points out a changing tendency of the uncertainty in convolutional layers and gives some insights to the interpretability of convolution.

The rest of this paper is organized as follows. Section \ref{related_work} discusses related works about adversarial learning and uncertainty. In Section \ref{section_3}, we propose the Min-Max property and give the theoretical proof. In Section \ref{section_4}, we discuss the relationship between the Min-Max property and uncertainty. Section \ref{section_5} experimentally analyzes the adversarial robustness of the Min-Max model and the standard model on MNIST and CIFAR10. Finally, Section \ref{section_6} concludes the paper.

\section{Related work} \label{related_work}

\subsection{Adversarial example}

Szegedy et al. \cite{r21_szegedy2013intriguing} first formulate the process of searching for adversarial examples as a box-constrained optimization problem:
\begin{eqnarray}
\begin{split}
&\min \|r\|_2 \\
&s.t. f(x+r)=l \\
& x+r\in [0,1]^m
\end{split}
\end{eqnarray}
where $f$ means a DNN model, and $r$ is a small perturbation. It is noted that adversarial examples are imperceptible to humans. This phenomenon raises great concern for the security of DNN models. Goodfellow et al. firstly propose a gradient-based attack, called Fast Gradient Sign Method (FGSM), which can fastly and effectively generate adversarial examples by computing the gradients only once. Moreover, they also try to explain the reasons for the existence of the adversarial examples, e.g., there are too many linear units in DNN models.
%where $f$ means a deep neural network, and $r$ is a small perturbation. It is noted that adversarial examples are imperceptible for human. This phenomenon raises great concerning for the security of deep neural networks. Googfellow et al. firsty propose a based-gradient attacks, called Fast Gradient Sign Method (FGSM), which can fastly and effectively generate adversarial examples by only computing gradients once time. Moreover, they also try to explain the existence of adversarial examples, e.g. there are too more linear units in deep neural networks.

%In general, all the attacks can be divide into two categories: white-box attacks and black-box attacks according to the knowledge of adversary. Under setting of white-box attacks, adversary can get all parameters and gradients from deep neural networks. However, under setting of black-box attacks, adversary cannot access to the trained deep neural networks but only can know the outputs of deep neural networks. It is obvious that attacks under black-box are harder than that under white-box. However, it is validated that transferability is a common property for adversarial examples \cite{r39_papernot2016transferability,r21_szegedy2013intriguing,r23_goodfellow2014explaining}. Papernot et al. \cite{r39_papernot2016transferability} showed that adversarial examples generated against a neural network can fool other neural network with same architectures. Therefore, attacks from white-box attacks can be applied to attack deep neural network under black-box setting through utilizing transferability.

According to the knowledge of the adversary, all the existing attacks can be divided into two categories: white-box attacks and black-box attacks. In white-box attacks, the adversary can get all the parameters and gradients from the DNN models. While in black-box attacks, the adversary cannot access the trained model but only knows its outputs. It is obvious that black-box attacks are more difficult to perform than white-box attacks. On the other hand, it is validated that transferability is a common property for adversarial examples \cite{r39_papernot2016transferability,r21_szegedy2013intriguing,r23_goodfellow2014explaining}. Papernot et al. \cite{r39_papernot2016transferability} showed that adversarial examples generated against a neural network can fool other neural networks with the same architecture. Therefore, white-box attacks can be applied to attack DNN models under a black-box setting by utilizing transferability.

The existing attack methods are briefly reviewed as follows. Carlini et al. \cite{r24_carlini2017towards} propose a non-gradient method, called CW, which constructs adversarial examples by optimizing an objective funciton chosen from several candidates, and found that adversarial training with adversarial examples generated by CW does not cause obfuscated gradients \cite{r26_athalye2018obfuscated}. Madry et al. \cite{r22_madry2017towards} propose a more powerful attack method based on projected gradient descent (PGD). They prove that PGD is the most powerful attack among all the first-order attacks. Besides, many other effective attacks have also been proposed \cite{r60_su2019one,r68_dong2018boosting,r50_kurakin2016adversarial,r69_kurakin2016adversarial,r27_moosavi2016deepfool,r28_papernot2016limitations}. Su et al. \cite{r60_su2019one} successfully attack DNNs by modifying only one pixel of the image. But under this attack, the pixel can be modified to any value. Moosavi-Dezfooli et al. \cite{r27_moosavi2016deepfool} propose a DeepFool by minimizing the distance between adversarial examples and decision boundaries. This distance can be used to measure the adversarial robustness of state-of-the-art DNN models. Papernet et al. \cite{r28_papernot2016limitations} propose a Jacobian-matrix-based attack, called Jacobian-based Saliency Map Attack. According to the saliency map of sample $x$, it is easy to find the pixel most sensitive to the change of $x$.

%Adversaral examples of above papers are imperceptible for human. But there are existed other type of adversarial examples, called adversarial patches. Brown et al. \cite{r70_brown2017adversarial} firstly introduce adversarial patch attacks for image classification. They generate adversarial patches by constructing a small patch with certain shape, e.g. circle or square, to stick to original image. This type of attack is widely used in object detection \cite{r71_liu2018dpatch,r72_karmon2018lavan,r74_eykholt2018robust,r73_sharif2016accessorize}.

Adversarial examples of the above papers are imperceptible to humans. But there is another type of adversarial examples, called adversarial patches. Brown et al. \cite{r70_brown2017adversarial} firstly introduce adversarial patch attacks for image classification. They generate adversarial patches by constructing a small patch with a certain shape, e.g. circle or square, to stick to the original image. This type of attack is widely used in object detection \cite{r71_liu2018dpatch,r72_karmon2018lavan,r74_eykholt2018robust,r73_sharif2016accessorize}.

\subsection{Adversarial robustness}

%Adversarial robustness refers to tolerance against adversarial examples. Adversarial examples are seriously threat for deep neural networks in security areas. Therefore, many works focus on this topic and proposed many effective defensive strategies against adversarial examples.
Adversarial robustness refers to the tolerance of the model against adversarial examples. Since adversarial examples are serious security threats to DNN models, many works try to find effective defense strategies against them.

%Distillation training \cite{r29_papernot2016distillation} is a effective way to improve adversarial robustness of model. Papernot et al. \cite{r29_papernot2016distillation} utilized two models with same architectures. One is trained to generate soft labels which are fed into other one. They found that this training schema can improve the adversarial robustness of model. However, this defense method fails against adversarial examples generated by CW.

Distillation training \cite{r29_papernot2016distillation} is an effective way to improve the adversarial robustness of the model. This training schema utilizes two models with the same architecture, one is trained to generate soft labels that are fed into the other one. This defense method is found to be effective in many scenarios but may fail against adversarial examples generated by CW.

%Adversarial training \cite{r21_szegedy2013intriguing,r50_kurakin2016adversarial,r35_zhang2019theoretically,r22_madry2017towards,r20_kurakin2018ensemble} is also a easy way to improving adversarial robustness of model. By including clean examples and adversarial examples, the model trained with these two types of examples are more adversarial robustness than standard training \cite{r21_szegedy2013intriguing,r22_madry2017towards}. Adversarial training at scale is not easy and very time-consuming. Prior attempts using adversarial training at ImageNet dataset were unsuccessful \cite{r50_kurakin2016adversarial}. But Tramer et al. \cite{r20_kurakin2018ensemble} utilize many of adversarial attacks to generate adversarial examples and propose ensemble adversarial training which augments training data with perturbations transferred from other models. Models trained with ensemble adversarial training are proved strong robustness to black-box attacks on ImageNet. Although adversarial training can improve adversarial robustness, some works \cite{r50_kurakin2016adversarial,r26_athalye2018obfuscated} have pointed out that adversarial training would cause obfuscated gradients. Athalye et al. \cite{r26_athalye2018obfuscated} point that obfuscated gradients is phenomenon what give a false sense of security against adversarial examples. They propose three types of obfuscated gradients and design respectively adversarial attacks which are successfully to attack all defensive papers in ICLR 2018 except for CW \cite{r24_carlini2017towards}.

Adversarial training \cite{r21_szegedy2013intriguing,r50_kurakin2016adversarial,r35_zhang2019theoretically,r22_madry2017towards,r20_kurakin2018ensemble} is another popular way to improve the adversarial robustness by including both clean examples and adversarial examples \cite{r21_szegedy2013intriguing,r22_madry2017towards}. Adversarial training at  scale is not easy and very time-consuming. Previous efforts made for adversarial training on the ImageNet dataset are unsuccessful \cite{r50_kurakin2016adversarial}. But Tramer et al. \cite{r20_kurakin2018ensemble} utilize many adversarial attacks to generate adversarial examples and propose ensemble adversarial training which augments training data with perturbations transferred from other models. Models by ensemble adversarial training have been proved to be robust to black-box attacks on ImageNet. Although adversarial training can improve adversarial robustness, some works \cite{r50_kurakin2016adversarial,r26_athalye2018obfuscated} have pointed out that it would cause obfuscated gradients. As pointed out by Athalye et al. \cite{r26_athalye2018obfuscated}, obfuscated gradient is a phenomenon that gives a false sense of security against adversarial examples. They propose three types of obfuscated gradients and design adversarial attacks that successfully attack all defensive papers in ICLR 2018 except that for CW \cite{r24_carlini2017towards}.

\subsection{Uncertainty in adversarial learning}
Basically, there are two kinds of uncertainties in a model.
\begin{enumerate}
\item Epistemic uncertainty, which is also called systematic uncertainty. This uncertainty is caused by lack of knowledge, i.e., the knowledge of the learner is not enough to choose a good model from the alternatives, thus the selected model may make inaccurate predictions.
\item Aleatoric uncertainty, which is also called statistical uncertianty. This uncertainty is produced during the model reasoning process, as a result, the model cannot provide an exact output but only a probability.
\end{enumerate}

%Entropy \cite{r75_lin1991divergence} is a straight-forward way to measure the uncertainty of model. Chen et al.\cite{r76_chen2019improving} propose a new loss function, called Guided Complement Entropy (GCE), which neutralizes the probability of the incorrect class o balance the correct class with the incorrect class. In order to be more confident about the predicted correct classes, they levelled off the probability of incorrect classes and made the distribution of incorrect classes as even as possible. They believe that more confidence in the prediction results can improve the accuracy of the prediction and also achieve the purpose of improving the adversarial robustness of model.

Entropy \cite{r75_lin1991divergence} is a straightforward way to measure the uncertainty of a model. Chen et al. \cite{r76_chen2019improving} propose an entropy-based loss function, called Guided Complement Entropy (GCE), which 
maximizes model's probabilities on the ground-truth class, and neutralizes model's probabilities on the incorrect classes to balance these two terms. They point out that GCE can improve adversarial robustness of a model, and at the same time, keep or improve the performace when no attack is present. Besides, training with GCE for models no longer needs adversarial examples.

%It is validated that the entropy of prediction confidence would become bigger after adversarial training \cite{r37_terzi2020directional}. Essentially, this idea just like to GCE. Smith et al. \cite{r63_smith2018understanding} study many of measure of uncertainty such as entropy and mutual information. In result, they shed light that mutual information seems to be effective at the task of adversarial detection. In fact, although deep neural networks achieve excellent performance, they often hard to capture their own uncertainties well. But Gaussian processes (GPs) \cite{r77_seeger2004gaussian} with RBF kernels on the other hand have better calibrated uncertainties and do not overconfidently predict. Considering these cases, Bradshaw et al. \cite{r78_bradshaw2017adversarial} propose GP hybrid deep networks, GPDNNs, which have better adversarial robustness against adversarial examples.

It is validated that the entropy of prediction confidence would become larger after adversarial training \cite{r37_terzi2020directional}, just like what happens for GCE. Except entropy, there exist many other measures of uncertainty in information theory. The relationship between different measures of uncertainty and adversarial robustness is worth further studying. Smith et al. \cite{r63_smith2018understanding} study many measures of uncertainty such as entropy and mutual information. As a result, they shed light that mutual information seems to be effective for the task of adversarial detection. Besides, the combination of neural networks and traditional machine learning methods is also helpful to analyze the uncertainty of a model. For example, Bradshaw et al. \cite{r67_bradshaw2017adversarial} point out that, although DNNs can achieve excellent performance, it is often difficult for them to capture their uncertainty well. Gaussian processes (GPs) \cite{r77_seeger2004gaussian} with RBF kernels on the other hand have better-calibrated uncertainties and do not overconfidently predict. Therefore, they propose a GP hybrid deep networks, GPDNNs, which have stronger adversarial robustness against adversarial examples.

%Uncertainty is a promising direction in adversarial learning. Uncertainty in adversarial learning is extending to some hot areas such as Object Detection \cite{r66_Liu_2019_ICCV} and reinforcement learning \cite{r79_pinto2017robust}. Liu et al. \cite{r66_Liu_2019_ICCV} utilize Monte Carlo sampling method to increase the model uncertainty. The proposed method, called Prior Driven Uncertainty Approximation (PD-UA), to generate a robust universal adversarial perturbation by fully exploiting the model uncertainty at each network layer. By formulating the policy learning as a zero-sum, e.g. an optimal destabilization policy, Pinto et al. \cite{r79_pinto2017robust} propose robust adversarial reinforcement learning (RARL). To model uncertainties, they train an agent which applies disturbance forces to the system.
Uncertainty modeling is a promising direction in adversarial learning, which has been used in some hot topics such as object detection \cite{r66_Liu_2019_ICCV} and reinforcement learning \cite{r79_pinto2017robust}. Liu et al. \cite{r66_Liu_2019_ICCV} utilize Monte Carlo sampling to increase model uncertainty. The proposed method, called Prior Driven Uncertainty Approximation (PD-UA), is to generate robust universal adversarial perturbation by fully exploiting the model uncertainty in each network layer. By formulating policy learning as a zero-sum problem, e.g., an optimal destabilization policy, Pinto et al. \cite{r79_pinto2017robust} propose robust adversarial reinforcement learning (RARL). To model uncertainties, they train an agent that applies disturbance forces to the system.
\section{Proposed method} \label{section_3}

\subsection{Framework of LeNet}
%LeNet \cite{r45_5265772} is a tiny neural network which includes two convolutional layers, two maxpooling layers and three fully connections layers. An overview of LeNet is described in Fig. \ref{fig:LeNet}. The non-linear activation function commonly is tanh \cite{r52_mishkin2015all}, or ReLU \cite{r46_glorot2011deep}, or ELU \cite{r53_clevert2015fast}. In this paper, we adopt ReLU as activation function since it was widely used \cite{r54_szegedy2016rethinking,r55_springenberg2014striving,r24_carlini2017towards,r29_papernot2016distillation}. It has a good performance when dealing some easy problems such as classification on MNIST \cite{r42_deng2012mnist} or FASHIION MNIST \cite{r44_xiao2017/online}. Due to the simple framework and fast training time, it is suitable to explore some features of neural network or design prototype of algorithm.

\begin{figure*}[ht]
	\centering

	\includegraphics[width=0.8\linewidth]{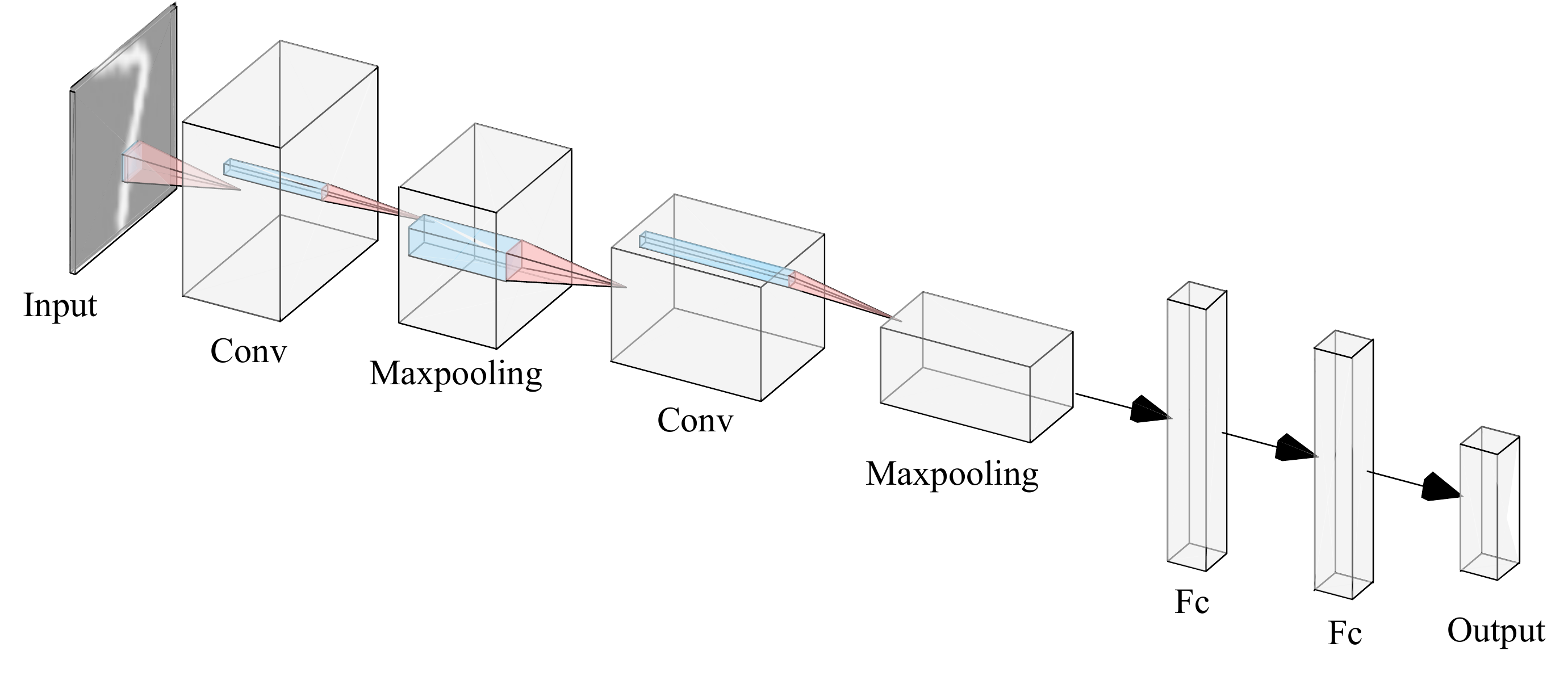}
	
	\caption{An overview of LeNet network \cite{r45_5265772}. This is a small neural network which only includes two convolution layers but make great achievement on some small dataset such as MNIST \cite{r42_deng2012mnist} and FASHION MNIST \cite{r44_xiao2017/online}.}
	\label{fig:LeNet}
\end{figure*}

LeNet \cite{r45_5265772} is a small neural network that includes two convolutional layers, two maxpooling layers, and three fully connected layers. An overview of LeNet is described in Fig. \ref{fig:LeNet}. The non-linear activation function can be tanh \cite{r52_mishkin2015all}, or ReLU \cite{r46_glorot2011deep}, or ELU \cite{r53_clevert2015fast}. In this paper, we adopt ReLU as the activation function since it is widely used in existing works \cite{r54_szegedy2016rethinking,r55_springenberg2014striving,r24_carlini2017towards,r29_papernot2016distillation}. LeNet has a good performance when dealing with simple classification problems such as MNIST \cite{r42_deng2012mnist} or FASHION MNIST \cite{r44_xiao2017/online}. Due to the simple structure and fast training speed, it can be used to explore features of the neural network or design prototype of the algorithm.

\subsection{Standard trainig and adversarial training}
\paragraph{Standard Training}
Given a training set with $N$ samples $D=\{(\boldsymbol{x}^{(i)}, y^{(i)}) \}^N_{i=1}$, we can learn a classifier $f:X \rightarrow Y$, where $\hat{y}=\mathop{\arg\max}_i f_i(\boldsymbol{x})$ represents the prediction result of the classifier on the sample $(\boldsymbol{x},y) \in \{(\boldsymbol{x}^{(i)}, y^{(i)})\}^N_{i=1}$. Then, the training process can be described as the following optimization problem:

\begin{equation}
\min \mathop{E}_{(\boldsymbol{x},y) \backsim D} [L(f(\boldsymbol{x}), y)] = \min \sum_{i=1}^{N} L(f(\boldsymbol{x}^{(i)}), y^{(i)}),
\label{eq-1}
\end{equation}
where $L$ represents the loss function.

\paragraph{Adversarial Training}
The set of adversarial examples concerning a sample $(\boldsymbol{x},y)$ is defined as 
\begin{equation}
\begin{split}
B_l(\boldsymbol{x},y,f)=\{\boldsymbol{x}_{adv} | \|\boldsymbol{x}_{adv}-\boldsymbol{x}\|_l \leq \epsilon ,\\ \arg\max_j f_j(\boldsymbol{x}_{adv}) \neq y\},
\end{split}
\end{equation}
where $l$ represents $l$-norm. Then, the process of its adversarial training can be formulated as follows:

\begin{equation}
\min \mathop{E}_{(\boldsymbol{x},y) \backsim D} [ \max_{\boldsymbol{x}_{adv} \in B_l} L(f(\boldsymbol{x}_{adv}), y)].
\label{eq-2}
\end{equation}

There are some differences between standard training and adversarial training as follows.

\begin{itemize}
	\item Adversarial training needs more samples than standard training. To improve the adversarial robustness of the model and make effective defense to both known and unknown attacks, the adversarial examples generated by powerful attacks would be added to the training dataset. The DNNs will learn more knowledge from adversarial examples than standard training.
	\item Adversarial training is more time-consuming than standard training. It is obvious that the adversarial examples are generated during the training process. The more complicated the generating algorithm, and the more iterations it includes, the longer time it will consume.
	\item Some works \cite{r20_kurakin2018ensemble,r26_athalye2018obfuscated,r47_qin2019adversarial} show that adversarial training may cause a phenomenon, called obfuscated gradients, which will give a false sense of security in defenses against adversarial examples. A good model should be able to make effective defense to adversarial examples, at the same time avoid obfuscated gradients.
	\item Adversarial training may decrease the accuracy of the model on a clean dataset. Tsipras et al. \cite{r51_tsipras2018robustness} point out that robustness may be conflicting with accuracy, causing the so-called accuracy-robustness problem. Empirically, many works \cite{r35_zhang2019theoretically,r49_su2018robustness,r50_kurakin2016adversarial,r9_wong2017provable,r38_raghunathan2018certified} show that adversarial training is to achieve a trade-off between adversarial robustness and accuracy.
\end{itemize}

\subsection{Empirical observation}

In order to analyse the difference between the standard model and the adversarial model, we make an investigation on the parameters of the convolutional layer. 

By using LeNet, we conduct two experiments for both standard training and adversarial training on MNIST dataset. The detailed setting such as learning rate, epochs and batchsize are listed in Section \ref{section_5}. It is noteworthy that, during adversarial training, the model updates parameters alternately between clean samples and adversarial samples.

\begin{figure*}[ht]
	\centering
	\subfloat[]{
		\includegraphics[width=0.45\linewidth]{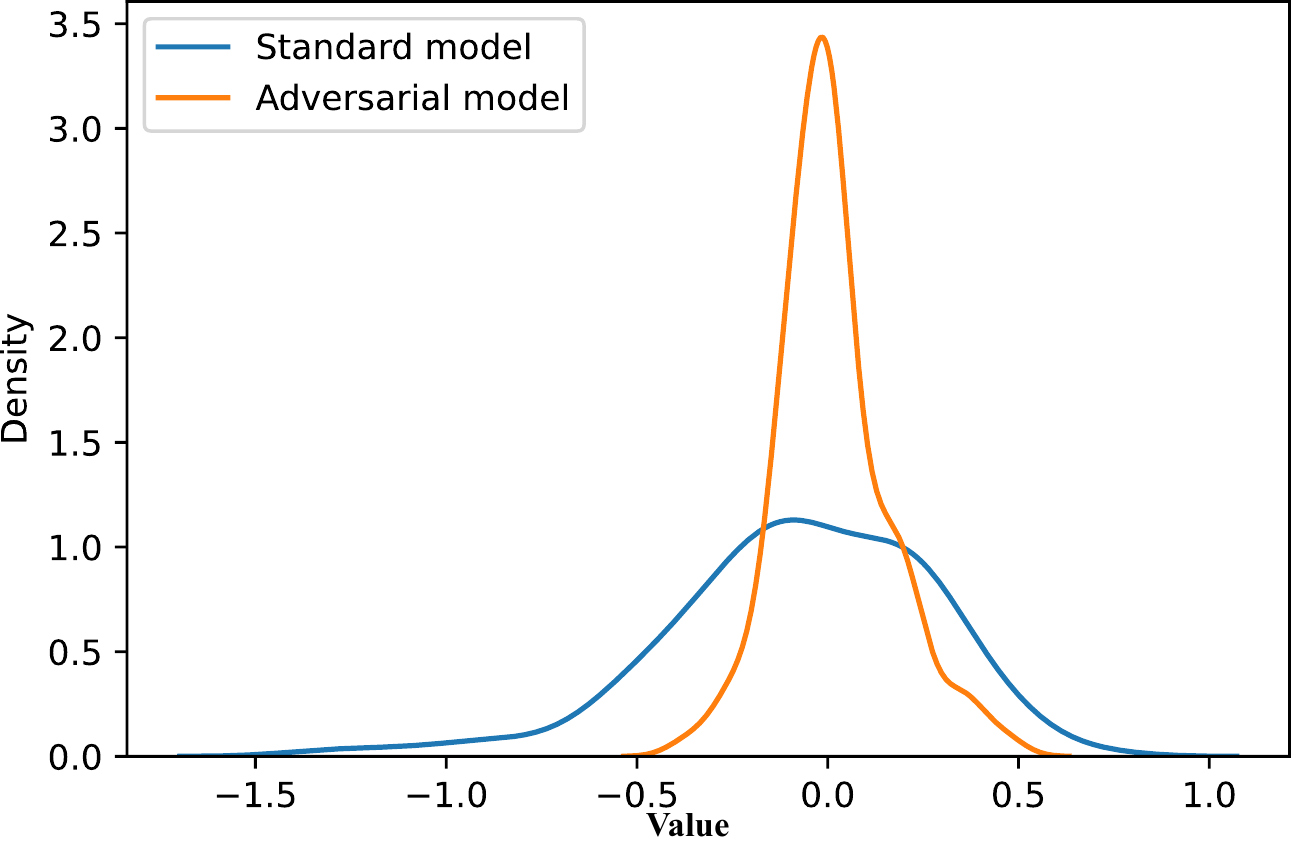}
	}
	\subfloat[]{
		\includegraphics[width=0.45\linewidth]{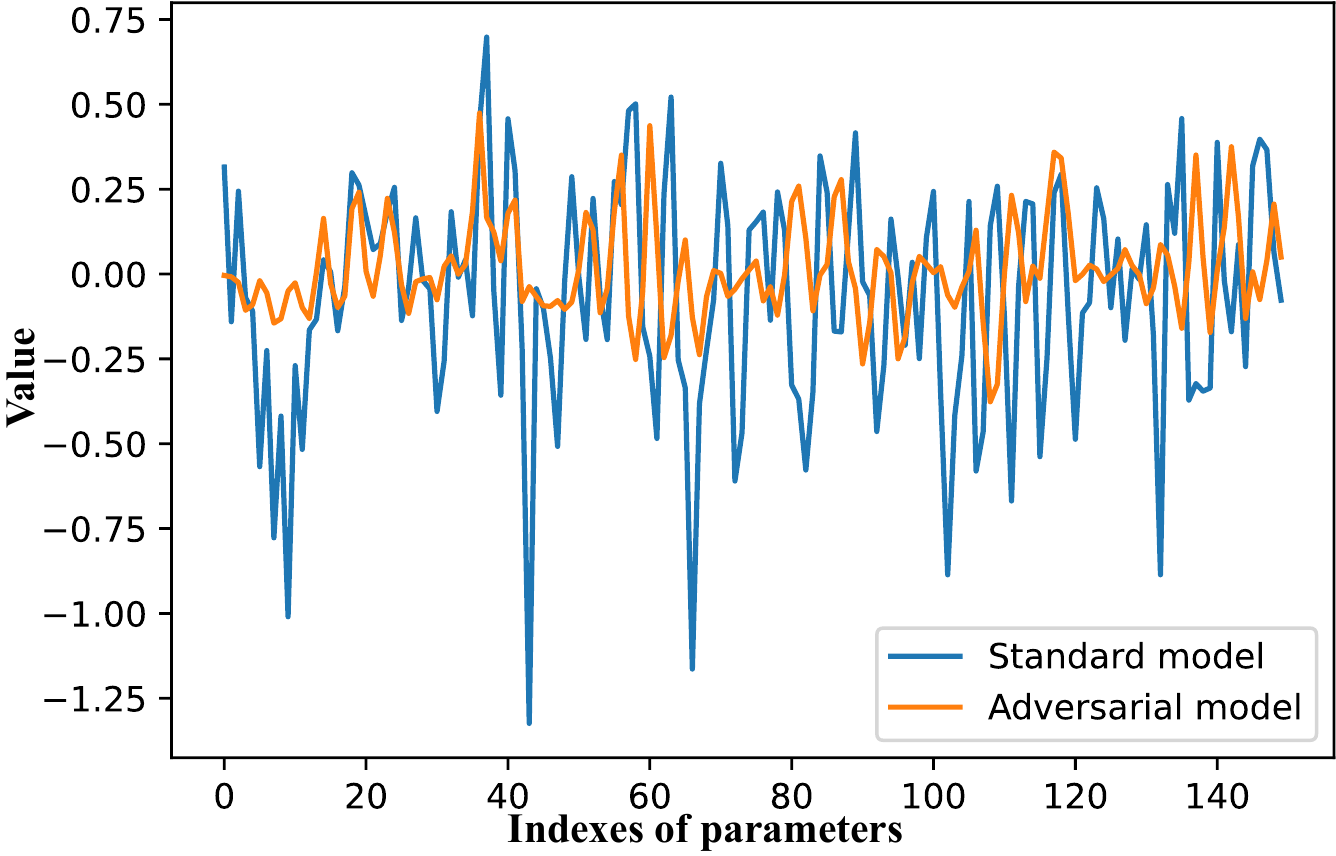}
	}
	\hfill
	\caption{Standard training and adversarial training: (a) The distribution of parameter of the first convolutional layer in LeNet. (b) The values of parameters of the first convolutional layer in LeNet network (Sort from left to right, top to bottom).}
	\label{fig:params}
\end{figure*}

Fig. \ref{fig:params} shows the distribution and the parameter values of the first convolutional layer in LeNet. It can be seen that after adversarial training, most parameters of the first convolutional layer become smaller, only a small part of them remain unchanged or become larger. By investigations on more data sets, we claim that this phenomenon is not accidential. It commonly exists in the first convolutional layer of LeNet, and sometimes can be found in the second convolutional layer.

Based on above discovery, we give the following proposition to explore more significant property in neural network.

\begin{proposition} \label{proposition}
	Given a neural network model, if most parameters in the convolutional layer approach zero, while only a small part of parameters are far away from zero, then the model have a good adversarial robustness. This property is called \textbf{Min-Max property}.
\end{proposition}

%%%%%%%%%%%%%%%%%%%%%%%%%%%%%%%%%%%%%
\subsection{Min-Max property}

%The robustness against adversarial examples for a deep model can be improved by adversarial training, which has been well acknowledged in the machine learning community and many domains of image processing. It is observed that, regarding the LetNet network, the absolute values of most weights in convolutional layers will become very small (approaching to zero) or very big (depending on the specific weights) after a several rounds of adversarial training. We call it Min-Max property which may be closely related to the several rounds of adversarial training.
The robustness of a DNN model against adversarial examples can be improved by adversarial training, which is well acknowledged in the machine learning and image processing communities. It is observed that in LeNet structure, the absolute values of most weights in convolutional layers will become very small (approaching zero) or very large (depending on the specific weights) after several rounds of adversarial training. We call this phenomenon Min-Max property.

It is noteworthy that a deep model with strong robustness against adversarial examples may not have the Min-Max property. However, based on a considerable number of simulations, it is observed that a deep model with the Min-Max property usually has strong robustness against adversarial examples. More importantly, the Min-Max property can be achieved without adversarial training. Thus, a well-trained deep model with the Min-Max property will possibly have strong robustness and adversarial training is no longer needed. Motivated by these observations, we propose a new training scheme as below based on Eq. \ref{eq-1}. 

\begin{equation}
\min \mathop{E}_{(\boldsymbol{x},y) \backsim D} [L(f(\boldsymbol{x}), y) + \lambda \|W_{conv}\|_1 - \mu\| W_{conv}\|^2_2],
\label{eq-3}
\end{equation}
where $\lambda$ and $\mu$ are hyperparamters, and $W_{conv}$ is the parameter of convolution. The term $\|W_{conv}\|_1$ makes the parameters approach zero, and while the term $\| W_{conv}\|^2_2$ makes the parameters be far away from zero.

In comparison with the original training scheme of Eq. \ref{eq-2}, Eq. \ref{eq-3} no longer needs adversarial training but still can improve the adversarial robustness of the model. In practice, we use the combination of $\lambda \| W_{conv}\|_1$ and $-\mu \| W_{conv}\|^2_2$ to control the Min-Max property of the model.

%In the following we try to mathematically explain why the Min-Max property is able to lead to strong robustness against adversarial examples via a simplified convolutional neural network.  Essentially convolutional operation is a linear transformation of feature space. Sometimes for simplification a RELU operation is required after convolution, which makes the result becoming piecewise linear. (Some scholars consider the convolution with RELU as a non-linear transformation in functionality but actually it is piecewise linear).  Considering the convolution as a linear transformation, we currently build two simplified models.
In the following, we try to mathematically explain why the Min-Max property is able to lead to strong robustness against adversarial examples via a simplified convolutional neural network. Essentially, convolutional operation is a linear transformation of feature space. For simplication, in some cases, a ReLU operation is required after convolution, which makes the result piecewise linear. It is noteworthy that some scholars consider the convolution with ReLU as a non-linear transformation in functionality but actually, it is piecewise linear. Based on these analysis, we propose two simplified models as follows.

Considering a single network $f(\boldsymbol{x})=\boldsymbol{W}\boldsymbol{x}$ where $\boldsymbol{W}$ is a $m\times n$ matrix, $\boldsymbol{x}$ is a $n\times 1$ vector, $m$ is the number of training samples and $n$ is the number of features. By adopting cross entropy loss, the loss function of this network is represented as $L(\boldsymbol{x};\boldsymbol{W})=-\boldsymbol{y}^T log(softmax(\boldsymbol{W}\boldsymbol{x}))$ where $\boldsymbol{y}$ is a one-hot vector, and 
\begin{equation*}
softmax(\boldsymbol{a})=exp(\boldsymbol{a})/(\boldsymbol{1}^T exp(\boldsymbol{a})).
\end{equation*}
The two models are given as follows:

\begin{itemize}
	\item $Model\#1$: $L_a(\boldsymbol{x};\boldsymbol{W_a})=-\boldsymbol{y}^T log(softmax(\boldsymbol{W_a}\boldsymbol{x}))$, where $W_a=(a_{ij})$ and $a_{ij}\sim U(0,1)$.
	\item $Model\#2$: $L_b(\boldsymbol{x};\boldsymbol{W_b})=-\boldsymbol{y}^T log(softmax(\boldsymbol{W_b}\boldsymbol{x}))$, where $W_b=(b_{ij})$ and $b_{ij}\sim B(1,p_1)$, $B(1,p_1)$ is the binomial distribution, and $p_1$ represents the probability of $b_{ij}=1$ where $p_1>0, p_1<<1$.
\end{itemize}

Suppose $a_{ij}$ follows a uniform distribution in [0, 1], and $b_{ij}$ follows a binomial distribution $B(1, p_1)$ where $p_1 << 1$ for each $i$ and $j$. We have the following theorem:

\begin{thm} \label{Theorem:1}
	Suppose $\boldsymbol{x}$ is a vector and $\boldsymbol{x}_i$ is the $i$th element in $\boldsymbol{x}$, $\forall i$, the following inequality holds
	$$\lvert \frac{\partial \mathbb{E}_{b_{ij}\sim B(1, p_1)} L_b(\boldsymbol{x};\boldsymbol{W_b})}{\partial \boldsymbol{x}_i}\rvert 
	\leq \lvert \frac{\partial \mathbb{E}_{a_{ij}\sim U(0,1)}L_a(\boldsymbol{x};\boldsymbol{W_a})}{\partial \boldsymbol{x}_i} \rvert.$$
\end{thm}

From Theorem \ref{Theorem:1} we know that the following inequality approximately holds:
\begin{equation*}
\begin{split}
\| L_b(\boldsymbol{x};\boldsymbol{W_b}) - L_b(\boldsymbol{x+\Delta};\boldsymbol{W_b})\| \leq \| L_a(\boldsymbol{x};\boldsymbol{W_a}) - \\ L_a(\boldsymbol{x+\Delta};\boldsymbol{W_a})\|,
\end{split}
\end{equation*}
where $\boldsymbol{\Delta}$ is a small perturbation of the observation $\boldsymbol{x}$. It shows that the perturbation of input to $Model\#1$ is more sensitive than that to $Model\#2$, which indicates that, to a great extent, the attacker can find the adversarial examples in the neighbourhood of clean samples for $Model\#1$ more easily than for $Model\#2$.

\begin{proof}
	Firstly, we compute the partial derivative $\frac{\partial L(\boldsymbol{x};\boldsymbol{W})}{\partial \boldsymbol{x}}$ where $L(\boldsymbol{x};\boldsymbol{W})=-\boldsymbol{y}^T log(softmax(\boldsymbol{W}\boldsymbol{x}))$. We mark $\boldsymbol{a}=\boldsymbol{W}\boldsymbol{x}$ where $\boldsymbol{W}$ is a $m\times n$ matrix and $\boldsymbol{x}$ is a $n \times 1$ vector. Let $\sigma(\boldsymbol{a})=softmax(\boldsymbol{a})$.
	
	Then, $L(\boldsymbol{x};\boldsymbol{W})$ can be rewritten in a scalar form:
	\begin{equation*}
	\begin{split}
	L(\boldsymbol{x};\boldsymbol{W}) &= -\boldsymbol{y}^T log(softmax(\boldsymbol{W}\boldsymbol{x})) \\
	&= -\boldsymbol{y}^T log\ \sigma(\boldsymbol{a}) \\
	&= -\sum_{j=1}^{m}y_j log\ \sigma_j(\boldsymbol{a}),
	\end{split}
	\end{equation*}
	where $\sigma_j(\boldsymbol{a})=\frac{\exp(a_j)}{\sum_{t=1}^{m} \exp(a_t)}$ and $a_j=\sum_{i=1}^{n}w_{ji}x_i$.
	
	We have
	\begin{equation*}
	\begin{split}
	\frac{\partial a_j}{\partial x_i} &= \frac{\partial}{\partial x_i} \sum_{k=1}^{m}w_{jk}x_k \\
	&= w_{ji}
	\end{split}
	\end{equation*}
	and
	\begin{equation*}
	\frac{\partial \sigma_i(\boldsymbol{a})}{\partial a_j} = \left\{
	\begin{aligned}
	\sigma_i (1-\sigma_j)	&	& {i=j} \\
	-\sigma_i \sigma_j		&	& {i\neq j}.
	\end{aligned}		
	\right.
	\end{equation*}		
	Applying the chain rule, we have
	\begin{equation*}
	\begin{split}
	\frac{\partial L}{\partial a_j} &= \sum_{i}\frac{\partial L}{\partial \sigma_i} \cdot \frac{\partial \sigma_i}{\partial a_j}=\frac{\partial L}{\partial \sigma_j}\cdot \frac{\partial \sigma_j}{\partial a_j}+\sum_{i\neq j} \frac{\partial L}{\partial \sigma_i}\cdot \frac{\partial \sigma_i}{\partial a_j} \\
	&= -y_j \frac{1}{\sigma_j}\cdot \sigma_j(1-\sigma_j) + \sum_{i\neq j}\frac{-y_i}{\sigma_i}\cdot (-1) \sigma_j \sigma_i \\
	&= -y_j(1-\sigma_j) + \sum_{i\neq j}y_i\sigma_j \\
	&= -y_j + y_j\sigma_j + \sum_{i\neq j}y_i\sigma_j \\
	&= \sigma_j - y_j.
	\end{split}
	\end{equation*}
	As a result we have
	\begin{equation}
	\begin{split}
	\frac{\partial L}{\partial x_i} &= \sum_{j=1}^{m} \frac{\partial L}{\partial a_j} \cdot \frac{\partial a_j}{\partial x_i} \\
	&= \sum_{j=1}^{m} w_{ji}\frac{\partial L}{\partial a_j} \\
	&= \sum_{j=1}^{m} w_{ji}(\sigma_j - y_j).
	\end{split} \label{eq:app:1}
	\end{equation}
	We can rewrite Eq. \ref{eq:app:1} in matrix form:
	\begin{equation*}
	\frac{\partial L(\boldsymbol{x};\boldsymbol{W})}{\partial \boldsymbol{x}}=\boldsymbol{W}^T(-\boldsymbol{y}+\sigma(\boldsymbol{W}\boldsymbol{x})).
	\end{equation*}
	Then
	\begin{equation*} \label{app:eq_1}
	\begin{split}
	&\lvert \frac{\partial \mathbb{E}_{a_{ij}\sim U(0,1)}[L_a(\boldsymbol{x};\boldsymbol{W_a})]}{\partial x_i} \rvert \\
	&= \lvert \mathbb{E}_{a_{ij}\sim U(0,1)}[\frac{\partial L_a(\boldsymbol{x};\boldsymbol{W_a})}{\partial x_i}] \rvert \\
	&= \lvert \mathbb{E}_{a_{ij}\sim U(0,1)}[\sum_{j=1}^{m} w_{ji}(\sigma_j - y_j)] \rvert \\
	&= \lvert \sum_{j=1}^{m} \frac{1}{2}(\sigma_j - y_j) \rvert \\
	&= \frac{1}{2}(1-\sum_{j=1}^{m}\sigma_j).
	\end{split}
	\end{equation*}
	In a similar way, we obtain
	\begin{equation*} \label{app:eq_2}
	\begin{split}
	&\lvert \frac{\partial \mathbb{E}_{b_{ij}\sim B(1,p_1)}[L_b(\boldsymbol{x};\boldsymbol{W_b})]}{\partial \boldsymbol{x_i}} \rvert
	= p_1(1-\sum_{j=1}^{m}\sigma_j).
	\end{split}
	\end{equation*}
	
	Since $p_1 > 0$ and $p_1 \ll (1-p_1)$, then $p_1 < \frac{1}{2}$. Therefore,
	$$ p_1(1-\sum_{j=1}^{m}\sigma_j) \leq \frac{1}{2}(1-\sum_{j=1}^{m}\sigma_j)$$
	Thus,
	$$\lvert \frac{\partial \mathbb{E}_{b_{ij}\sim B(1, p_1)} L_b(\boldsymbol{x};\boldsymbol{W_b})}{\partial \boldsymbol{x}_i}\rvert \\
	\leq \lvert \frac{\partial \mathbb{E}_{a_{ij}\sim U(0,1)}[L_a(\boldsymbol{x};\boldsymbol{W_a})]}{\partial \boldsymbol{x}_i} \rvert$$
	If $p_1=\frac{1}{2}$ then the equality holds.
	Thus, Theorem \ref{Theorem:1} is proven. $\hfill\blacksquare$
\end{proof}

\section{Uncertainty in Min-Max property} \label{section_4}

In this section, we try to explain the Min-Max property from the perspective of uncertainty. It has been proved that uncertainty is inevitable when making decision for a sample \cite{r57_yeung2002improving}. This uncertainty is usually caused by the fuzziness of the similarity matrix.

In the proposed training scheme, i.e., Eq. \ref{eq-3}, the minimization of $(\|W_{conv}\|_1 - \|W_{conv}\|^2_2)$ makes the parameters have the Min-Max property. For the convenience of computing, let
\begin{eqnarray}
	M(W_{conv}) = \|W_{conv}\|_1 - \|W_{conv}\|_2^2,
	\label{eq-4}
\end{eqnarray}
where we consider $W_{conv}$ as a vector denoted by
\begin{eqnarray}
	W_{conv} = [w_1, w_2, ..., w_n]^T.
\end{eqnarray}

In view of the changing tendency of $w_i\ (1\leq i \leq n)$, minimizing Eq. \ref{eq-4} equals to minimizing $M(W_{conv})$. 

Then we have
\begin{equation}
\begin{split}
M(W_{conv}) &= \|W_{conv}\|_1 - \|W_{conv}\|_2^2 \\
&= \sum_{i=1}^{n}|w_i| - \sum_{1}^{n}|w_i|^2 \\
&= \sum_{i=1}^{n} |w_i|(1-|w_i|).
\end{split}
\label{eq-5}
\end{equation}
Eq. \ref{eq-5} is derived from a simple function, i.e.,
\begin{eqnarray}
f(x)=|x|(1-|x|), x\in R.
\end{eqnarray}
From this function, we have
\begin{equation}
f^\prime (x)=\left\{
\begin{aligned}
1-2x	&	,& {x\geq 0} \\
-1-2x	&	,& {x<0}.
\end{aligned}		
\right.
\end{equation}
Thus, 
\begin{equation}
\left\{
\begin{aligned}
f^\prime (x) \geq 0	&	,& {x\leq -\frac{1}{2} \cup 0\leq x \leq \frac{1}{2}} \\
f^\prime (x) < 0	&	,& {-\frac{1}{2}<x<0 \cup x>\frac{1}{2}}.
\end{aligned}		
\right.
\end{equation}

Therefore, from Eq. \ref{eq-4}, we know that $M$ with respect to $w_i$ is a strictly monotonically increasing function under the condition of $x\leq -\frac{1}{2}\cup 0\leq x \leq \frac{1}{2}$ and is a strictly monotonically decreasing function under the condition of $-\frac{1}{2}<x<0\cup x>\frac{1}{2}$. By minimizing Eq. \ref{eq-4}, the positive values in the parameters will be far away from $0.5$ and the negative values in the parameters will be far away from $-0.5$. Based on this statement, we have the following equality
\begin{eqnarray}
\begin{split}
\min M(W_{conv}) = \lim\limits_{w_i \rightarrow 0 \ \text{or}\  w_i \rightarrow \infty} M(W_{conv}).
\end{split}
\end{eqnarray}

According to \cite{r58_basak1998unsupervised}, the fuzziness of this vector can be define as
\begin{eqnarray}
\begin{split}
Fuzziness(W_{conv}) = - \frac{1}{N}\sum_{1}^{N}[\rho_i log \rho_i + \\ (1-\rho_i) log(1-\rho_i)],
\end{split}
\label{eq-6}
\end{eqnarray}
where $N$ is the size of vector $W_{conv}$ and $\rho_i$ is defined as follows:
\begin{eqnarray}
\rho_{i} = \frac{1}{1+|w_i|}.
\label{eq-7}
\end{eqnarray}
From Eq. \ref{eq-7}, we have
$$w_i \rightarrow 0 \Rightarrow \rho_{i} \rightarrow 1,$$
$$w_i \rightarrow \infty \Rightarrow \rho_{i} \rightarrow 0.$$
These two cases (e.g. $\rho_{i} \rightarrow 1$ and $\rho_{i} \rightarrow 0$) are far away from $0.5$ which is the extreme value of Eq. \ref{eq-6}. Therefore, the fuzziness of the parameters will decrease as minmizing the $M(W_{conv})$.

\section{Experiments and results} \label{section_5}

In this section, we evaluate the adversarial robustness of the proposed model on MNIST \cite{r42_deng2012mnist} dataset and CIFAR10 \cite{r43_krizhevsky2009learning} dataset by adopting LeNet as the network structure. The adversarial robustness of the model is measured by the accuracy under adversarial attacks. Besides, we also calculate the fuzziness of parameters in convolutional layer to evaluate the uncertainty of the model.
\subsection{Experimental setup}

%We conduct experiments with Python 3.6 and utilize toolbox, called \textit{Advertorch} \cite{r56_ding2019advertorch}, to evalute the adversarial robustness of model. And before training, the input values are normalized by $(x-\mu)/\sigma$ where $\mu$ and $\sigma$ are the mean and variance among all samples. Standard model mean a model trained with clean examples and cross entropy. Min-Max model mean a model trained with clean examples, adversarial examples and our objective function designed according Min-Max property. Adv. model mean a model trained with clean examples and adversarial examples.

The experiments are conducted under Python 3.6 with the Advertorch toolbox \cite{r56_ding2019advertorch}. Before training, the input values are normalized by $(x-\mu)/\sigma$, where $\mu$ and $\sigma$ are the mean and variance among all samples. For MNIST, we have $\mu=0.1307$ and $\sigma=0.3081$; while for CIFAR10, we have $\mu=(0.4914, 0.4822, 0.4465)$ and $\sigma=(0.2023, 0.1994, 0.2010)$.

We compare the performance of four models without adversarial training:
\begin{itemize}
	\item Standard model (Std.): This is a model trained with clean examples by adopting cross-entropy as the loss function.
	\item Standard model with L1 normalization (Std.+L1): This is a model trained with clean examples, and the loss function includes an additional L1 normalization related to parameters.
	\item Standard model with L2 normalization (Std.+L2): This is a model trained with clean examples, and the loss function includes and additional L2 normalization related to parameters.
	\item Min-Max model (Min-Max): This is a model trained with clean examples and our proposed objective function \ref{eq-3}.
\end{itemize}
Std.+L1 and Std.+L2 are designed for ablation study since the proposed objective function is composed of L1 normalization and L2 normalization. Moreover, these models have the same framework and setting.

As for the datasets, we briefly introduce them as follows:
\begin{itemize}
	\item MNIST dataset consists of 60,000 training samples and 10,000 testing samples, each of which is a 28x28 pixel handwriting digital image. By setting $batchsize=64, lr=0.001, epochs=10$, we use \textit{Adam} optimization to train the neural network. When evaluating the model with Advertorch, we adopt some attacks (FGSM \cite{r23_goodfellow2014explaining}, PGD \cite{r22_madry2017towards}, CW \cite{r24_carlini2017towards}, MIA \cite{r68_dong2018boosting}, L2BIA \cite{r50_kurakin2016adversarial} and LinfBIA \cite{r50_kurakin2016adversarial}) widely used where $epsilon=0.3$.
	\item CIFAR10 dataset is composed of 60,000 32x32 colour image, 50,000 for training and 10,000 for testing. By setting $batchsize=64, lr=0.01, epochs=90$, we use \textit{Adam} optimization to train the neural network where the learning rate halved every 30 epochs. When evaluating the adversarial robustness, we set $epsilon=0.03$.
\end{itemize}

\subsection{Adversarial robustness}
We use six attack methods to evaluate the adversarial robustness of the model, i.e., FGSM \cite{r23_goodfellow2014explaining}, PGD \cite{r22_madry2017towards}, CW \cite{r24_carlini2017towards}, MIA \cite{r68_dong2018boosting}, L2BIA \cite{r50_kurakin2016adversarial} and LinfBIA \cite{r50_kurakin2016adversarial}. The higher accuracy under adversarial attack, the stronger adversarial robustness of the model. We set $epsilon=0.3$ for MNIST dataset and set $epsilon=0.03$ for CIFAR10 dataset.

\begin{figure}[ht]
	\centering
	\includegraphics[width=0.95\linewidth]{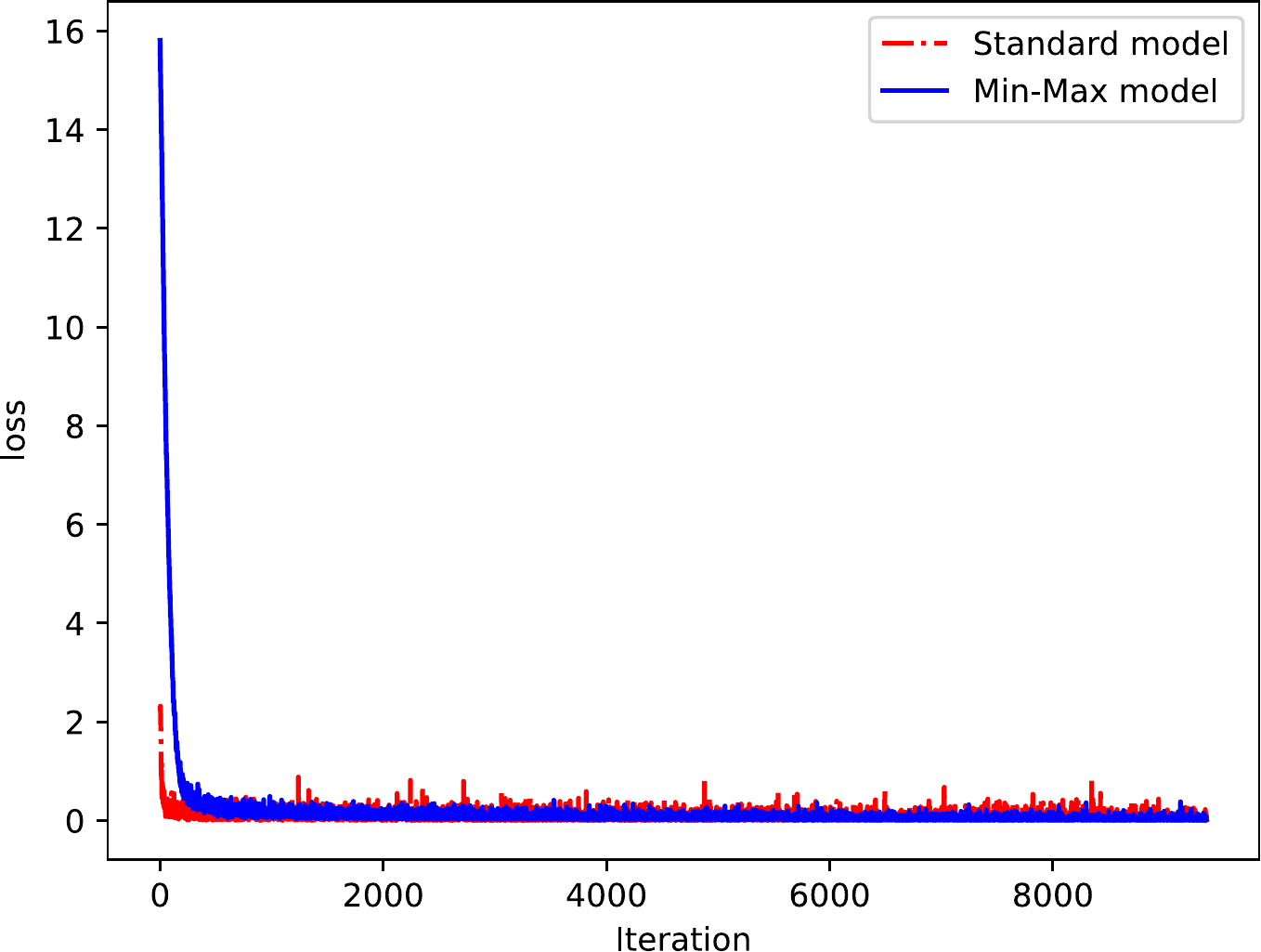}
	\caption{MNIST: The comparison of training loss between standard training and adversarial training. After convergence, Min-Max model is more stable than standard model.}
	\label{fig:MNIST_loss}
\end{figure}

%Table \ref{tab:MNIST-1} shows adversarial robustness of standard model and Min-Max model by testing the accuracy on test dataset under various adversarial attacks. It can be seen that our model, Min-Max model, has better adversarial robustness than standard model under all of attack methods. It experimentally verifies that minimizing Eq. \ref{eq-3} can make assign Min-Max property to the model. And model with Min-Max property have a better adversarial robustness. Then it experimentaly verifies the correctness of Theorem \ref{Theorem:1}. Moreover, from Fig. \ref{fig:MNIST_loss} which shows the training loss of standard model and Min-Max model, we find that Min-Max model converges slower than standard model but more stable after convergence. 
Table \ref{tab:MNIST-1} shows the adversarial robustness of the standard model and Min-Max model regarding the testing accuracy under various adversarial attacks. It can be seen that the proposed the Min-Max model has stronger adversarial robustness than the standard model under all of the attack methods. This result experimentally verifies the correctness of Theorem \ref{Theorem:1}, i.e., the model will have the Min-Max property by minimizing Eq. 5, and this property makes it have stronger adversarial robustness. Moreover, Fig. \ref{fig:MNIST_loss} shows the training loss of the standard model and Min-Max model. It can be observed that the Min-Max model converges slower than the standard model but is more stable after convergence.

%\begin{table}[ht]
%	\caption{MNIST: The comparison of accuracy between standard model and Min-Max model. Standard model have the same framework and setting with Min-Max model. But the loss function of standard model is cross entropy and Min-Max model is \ref{eq-3}.}
%	\centering
%	\label{tab:MNIST-1}
%	\begin{tabular}{lll}
%		\toprule
%		& Standard model & Min-Max model \\ \midrule
%		No attack & \textbf{0.9861}         & 0.9844        \\
%		FGSM      & 0.8152         & \textbf{0.9475}        \\
%		PGD       & 0.5928         & \textbf{0.8369}        \\
%		CW        & 0.5697         & \textbf{0.6892}        \\
%		MIA       & 0.6268         & \textbf{0.6980}       \\ 
%		L2BIA     & 0.9747         & \textbf{0.9765}   \\
%		LinfBIA   & 0.6046         & \textbf{0.8599}   \\
%		JSMA      & 0.9946         & \textbf{0.9981} \\ \bottomrule
%	\end{tabular}
%\end{table}

\begin{table}[ht]
	\caption{MNIST: The comparison of accuracy between standard models (Std., Std.+L1 and Std.+L2) and Min-Max model.}
	\centering
	\label{tab:MNIST-1}
	\begin{tabular}{lllll}
		\toprule
		& Std.  & Std.+L1 & Std.+L2         & Min-Max  \\ \midrule
		No attack & \textbf{0.9861} & 0.9747  & 0.9576          & 0.9844          \\
		FGSM      & 0.8152          & 0.7610  & 0.7631          & \textbf{0.9475} \\
		PGD       & 0.5928          & 0.6198  & 0.7130          & \textbf{0.8369} \\
		CW        & 0.5697          & 0.4961  & 0.5403          & \textbf{0.6892} \\
		MIA       & 0.6268          & 0.6190  & \textbf{0.7181} & 0.6980          \\
		L2BIA     & 0.9747          & 0.9659  & 0.9501          & \textbf{0.9765} \\
		LinfBIA   & 0.6046          & 0.6276  & 0.7126          & \textbf{0.8599} \\ \bottomrule
	\end{tabular}
\end{table}

Table \ref{tab:CIFAR-1} further reports the comparison results between the standard model and the Min-Max model on a more complicated dataset, i.e., CIFAR10.
It can be seen that the Min-Max property significantly improves the adversarial robustness of the model, which is consistent with the conclusion on MNIST dataset. However, this improvement is limited. For some powerful adversarial attacks such as CW, the model is still vulnerable. This may be due to the fact that CW is a non-gradient adversarial attack. Min-Max property in the convolutional layer can avoid sharp change by small perturbation which is spread by the gradient in the neural network. However, if the small perturbation is not dependent on the gradient, it will be difficult for the Min-Max property to catch the exception.

It is noteworthy that the proposed objective function, i.e., Eq. \ref{eq-3}, contains both L1 normalization and L2 normalization. In order to investigate the effectiveness of these two normalizations, we ablate one of them.
As shown in Tables \ref{tab:MNIST-1} and \ref{tab:CIFAR-1}, L1 normalization does not help to improve the adversarial robustness and L2 normalization can improve the adversarial robustness a little bit. However, by combining L1 normalization and L2 normalization, the Min-Max model has much stronger adversarial robustness than both Std.+L1 and Std.+L2 under most attacks. This shows that Eq. \ref{eq-3} can effectively make the parameters either closer to zero or further away from zero, as described in the Min-Max property.
\begin{table}[ht]
	\centering
	\caption{CIFAR10: The comparison of accuracy between standard models (Std., Std.+L1 and Std.+L2) and Min-Max model.}
	\label{tab:CIFAR-1}
	\begin{tabular}{lllll}
		\toprule
		& Std. & Std.+L1 & Std.+L2         & Min-Max   \\ \midrule
		No attack & 0.6352         & 0.5589  & \textbf{0.6573} & 0.5668          \\
		FGSM      & 0.2382         & 0.3144  & 0.3716          & \textbf{0.3927} \\
		PGD       & 0.1669         & 0.2937  & 0.3416          & \textbf{0.3712} \\
		CW        & 0.0000         & 0.0000  & 0.0000          & 0.0000          \\
		MIA       & 0.1583         & 0.2918  & 0.3382          & \textbf{0.3633} \\
		L2BIA     & 0.6249         & 0.5550  & 0.5561          & \textbf{0.6512} \\
		LinfBIA   & 0.2128         & 0.3135  & 0.3594          & \textbf{0.3860} \\ \bottomrule
	\end{tabular}
\end{table}

\subsection{Uncertainty}
We make an investigation on the fuzziness of parameters in the first convolutional layer of LeNet using Eq. 14. The standard model, the adversarial model and the Min-Max model are compared. It is noteworthy that the adversarial model is trained alternately between clean examples and adversarial examples in each epoch, where the adversarial examples are generated by PGD \cite{r22_madry2017towards}. With this training scheme, the phenomenon of accuracy-robustness problem would reduce.

\begin{figure}[ht]
	\centering
	\includegraphics[width=0.95\linewidth]{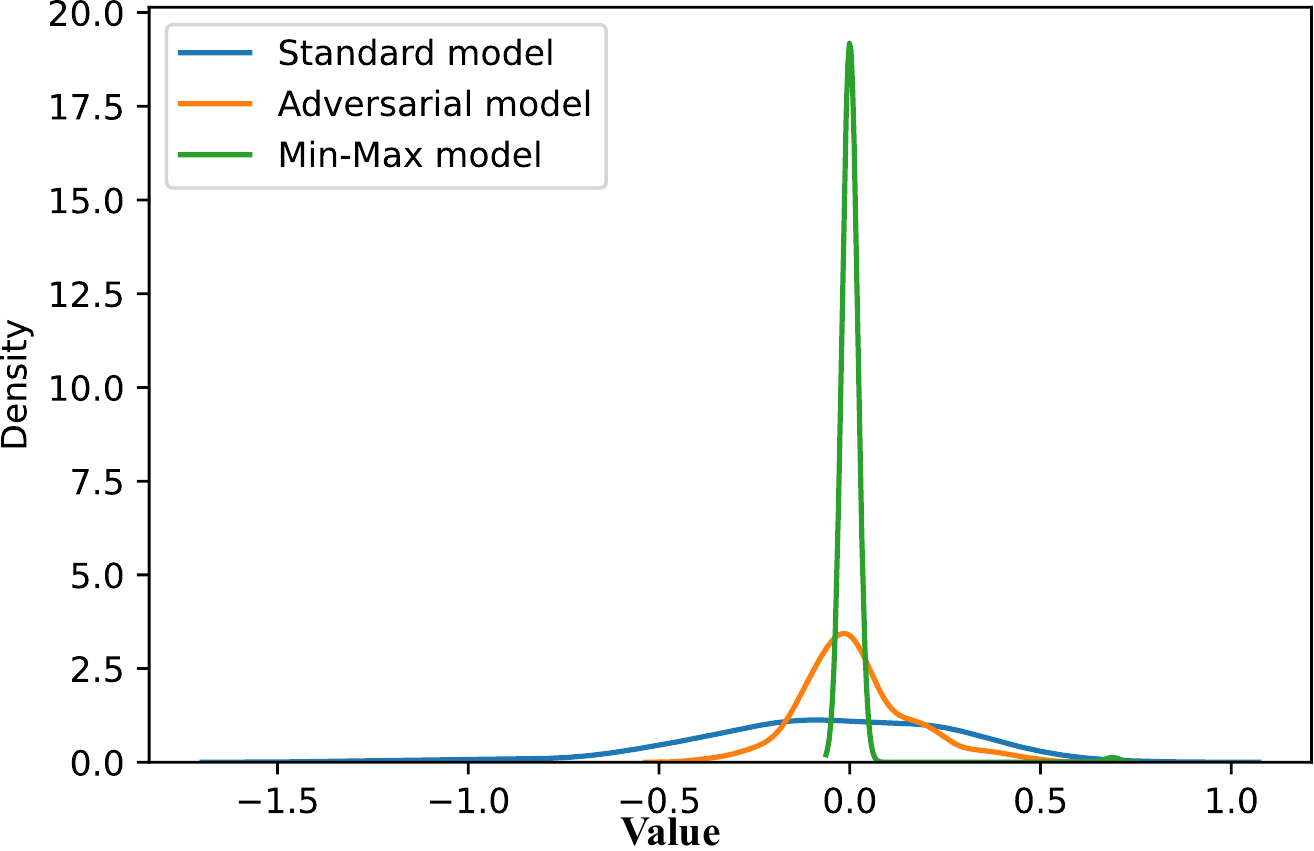}
	\caption{MNIST: The comparison of parameter distribution in the first convolutional layer of LeNet among standard model, adversarial model and Min-Max model.}
	\label{fig:fuzziness_MNIST}
\end{figure}

\begin{figure}[ht]
	\centering
	\includegraphics[width=0.95\linewidth]{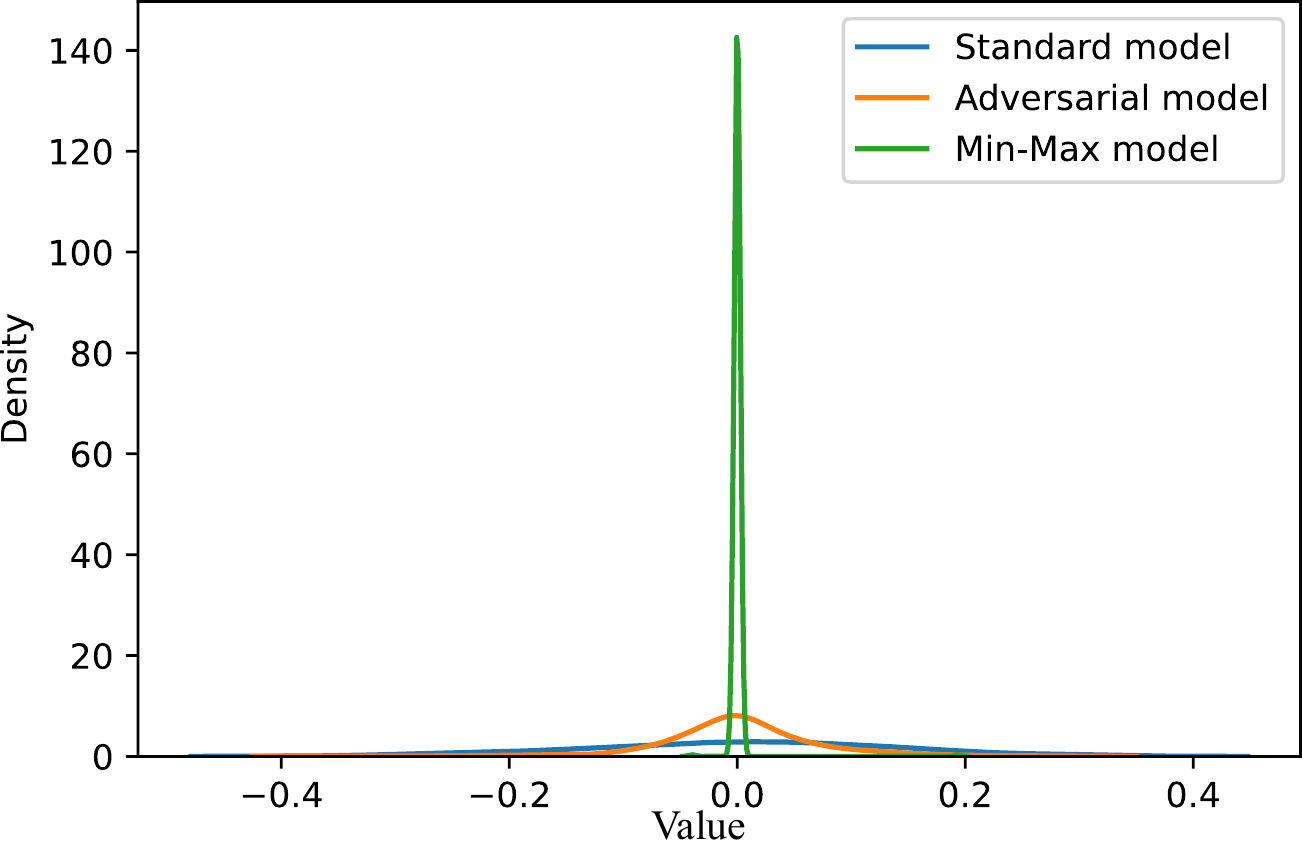}
	\caption{CIFAR10: The comparison of parameter distribution in first convolutional layer of LeNet among standard model, adversarial model and Min-Max model.}
	\label{fig:fuzziness_cifar}
\end{figure}

Figs. \ref{fig:fuzziness_MNIST} and \ref{fig:fuzziness_cifar} show the distribution of parameters in the first convolutional layer of LeNet for dataset MNIST and CIFAR10, respectively. It can be seen that for both of these two datasets, the number of parameters close to zero in the adversarial model is larger than that in the standard model. This is the basis for the proposed Min-Max property. Then, by further using Eq. 5, this property in the Min-Max model becomes more obvious. The intuitive result is shown in Figs. \ref{fig:fuzziness_MNIST} and \ref{fig:fuzziness_cifar}. The number of parameters close to zero in the Min-Max model is about 17-19 times of that in the standard model for MNIST, and is about 140-150 times for CIFAR10. We suspect that this extreme phenomenon is responsible for the decline in the accuracy on clean examples, as a result, the accuracy of the Min-Max model on clean examples is lower than that of the standard model. 

Moreover, we utilize fuzziness (Eq. \ref{eq-6}) to quantify the Min-Max property. As shown in Table \ref{tab:fuzziness}, the fuzziness of the adversarial model is lower than that of the standard model, demonstrating a lower uncertainty of the adversarial model. By minimizing Eq. \ref{eq-3}, the uncertainty of the Min-Max model is further reduced. 

\begin{table}[ht]
	\centering
	\caption{The fuzziness of parameters of fist convolutional layer in LeNet.}
	\label{tab:fuzziness}
	\begin{tabular}{lll}
		\toprule
		& MNIST  & CIFAR10 \\ \midrule
		Standard model & 0.4417 & 0.2842  \\
		Adv. model     & 0.2611 & 0.1594  \\
		Min-Max model  & \textbf{0.0053} & \textbf{0.0016}  \\ \bottomrule
	\end{tabular}
\end{table}

\begin{figure*}[ht]
	\centering
	\subfloat[]{
		\includegraphics[width=0.9\linewidth]{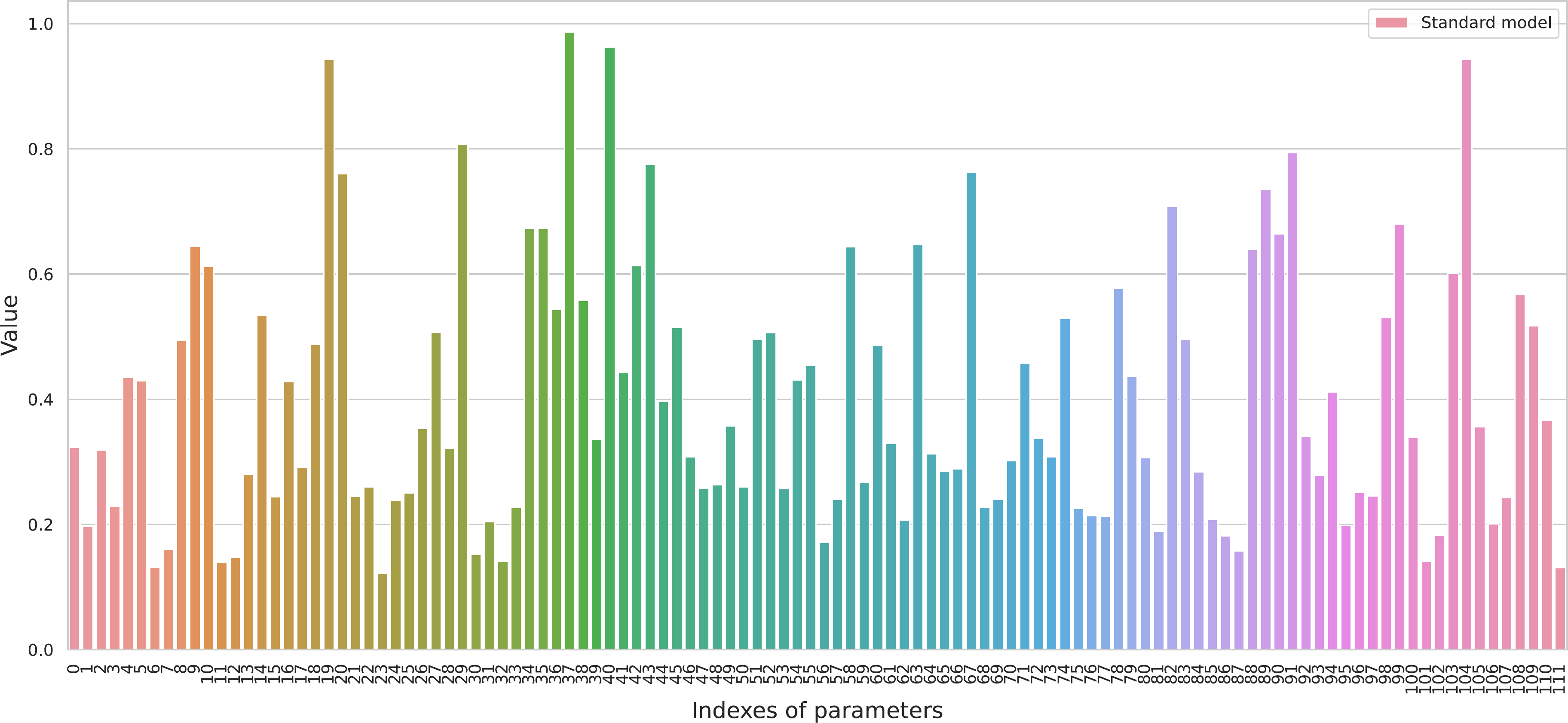}
	}
	\hfill
	\subfloat[]{
		\includegraphics[width=0.9\linewidth]{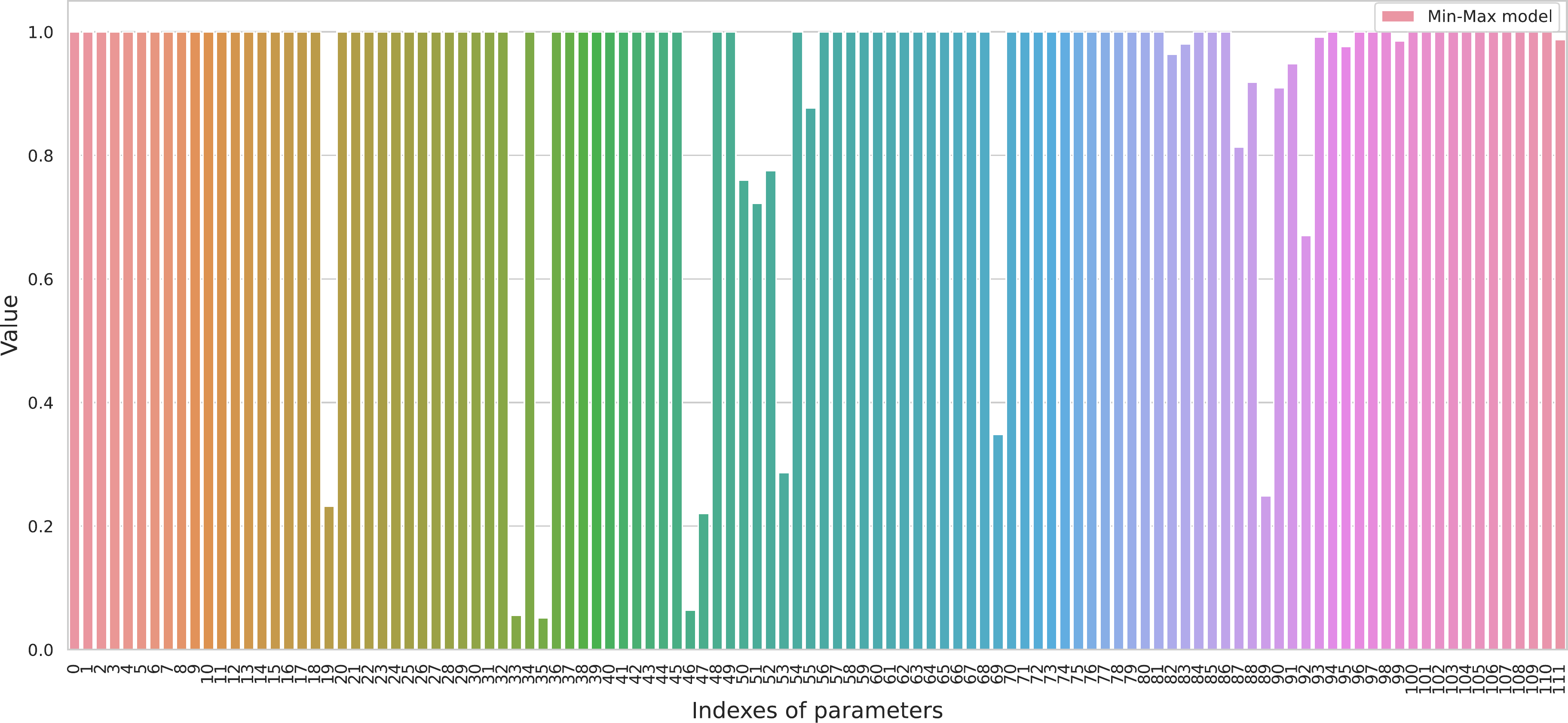}
	}
	\hfill
	\caption{The part of normalized parameters in first convolutional layer in the task of CIFAR10: (a) The normalized parameters in standard model. (b) The normalized parameters in Min-Max model.}
	\label{fig:uncertainty_cifar_show}
\end{figure*}

\subsection{Disscussion}

It is investigated that Min-Max property would cause a drop in accuracy on clean examples. This phenomenon is not obvious on MNIST but is easily aware on CIFAR10. As shown in Table \ref{tab:CIFAR-1}, accuracy of the Min-Max model on clean examples drops about 0.089 compared with the standard model. This phenomenon brings the so-called accuracy-robustness problem. Although the trade-off between accuracy and robustness is widely discussed \cite{r51_tsipras2018robustness,r35_zhang2019theoretically,r49_su2018robustness,r50_kurakin2016adversarial,r9_wong2017provable,r38_raghunathan2018certified}, the underlying theories still remain unknown. Through observing the quantitative indicator of the uncertainty, e.g., fuzziness, we find that the uncertainty of the Min-Max model has dropped to be too low as shown in Table \ref{tab:fuzziness}. Similar observations can also be made from Figs. \ref{fig:fuzziness_MNIST} and \ref{fig:fuzziness_cifar}. Therefore, we guess that the accuracy-robustness problem is caused by this extreme phenomenon of the Min-Max property to some extent. 

%Finally, we observe the normalized convolutional vector in both standard training and our training strategies. We show part of the parameters in the first convolutional layer for CIFAR10 in Fig. \ref{fig:uncertainty_cifar_show} (为啥文中图4的引用出现在图5跟图6后边？), where the values are ranked according to their indexes of positions, and are normalized into a fixed range [0, 1] by Eq. \ref{eq-7}. It is easy to see that the number of values close to zero or close to one in the Min-Max model are much larger than that in the standard model. This intuitively shows that our training strategies have a much lower uncertainty than the standard model. 

Finally, we observe the normalized convolutional vector in both standard training and our training strategies. We show part of parameters in the first convolutional layer for the task of CIFAR10, where the values are ranked according to their indexes of position. By Eq. \ref{eq-7}, we normalize the values into a fixed range ([0, 1]). As shown in Fig. \ref{fig:uncertainty_cifar_show}, it is easy to see that the numbers approach zero or one in the Min-Max model are much larger than that of the standard model. It intuitively shows that the uncertainty of our training strategies is far lower than that of the standard model.

\section{Conclusion} \label{section_6}

The Min-Max property is extracted by observing the difference between the standard model and the adversarial model in the convolutional layer. This Min-Max property is a phenomenon that the parameters in the convolutional layer will be either closer to zero or farther away from zero. Possessing the Min-Max property of the model have stronger  adversarial robustness. According to this property, we design a new objective function. And minimizing this objective function equals making the model with the Min-Max property. We theoretically and experimentally validate the correctness of the Min-Max property. Moreover, from the perspective of uncertainty, we use fuzziness to quantify the Min-Max property. As a result, the process of maximizing the Min-Max property is equivalent to minimizing the fuzziness of parameters. In future works, we will explore the Min-Max property in a more complicated DNN structure and try to discover properties of uncertainty in DNN. Moreover, how to further improve the adversarial robustness based on uncertainty analysis is also worth to study in the future. 

\begin{acknowledgements}
This work was supported in part by Natural Science Foundation of China (Grants 61732011 and 61976141,61772344), in part by the Natural Science Foundation of SZU (827-000230), and in part by the Interdisciplinary Innovation Team of Shenzhen University.
%in part by the National Natural Science Foundation of China (Grants 61976141, 61772344 and 61732011), in part by the Natural Science Foundation of SZU (827-000230), and in part by the Interdisciplinary Innovation Team of Shenzhen University.
\end{acknowledgements}

% Authors must disclose all relationships or interests that 
% could have direct or potential influence or impart bias on 
% the work: 
%
\section*{Conflict of interest}
The authors declare that there is no conflict of interests regarding the publication of this article.
%
% The authors declare that they have no conflict of interest.

% BibTeX users please use one of
%\bibliographystyle{spbasic}      % basic style, author-year citations
\bibliographystyle{spmpsci}      % mathematics and physical sciences
\bibliography{reference}   % name your BibTeX data base

% Non-BibTeX users please use
%\begin{thebibliography}{}
%%
%% and use \bibitem to create references. Consult the Instructions
%% for authors for reference list style.
%%
%\bibitem{RefJ}
%% Format for Journal Reference
%Author, Article title, Journal, Volume, page numbers (year)
%% Format for books
%\bibitem{RefB}
%Author, Book title, page numbers. Publisher, place (year)
%% etc
%\end{thebibliography}

\end{document}